\documentclass[sigconf]{acmart}
\usepackage{graphicx}
\usepackage{subfig}
\usepackage{multirow}
\usepackage{tabularx}
\usepackage{array}

\usepackage{lipsum}
\AtBeginDocument{%
	}

\setcopyright{acmlicensed}
\copyrightyear{2025}
\acmYear{2025}
\acmDOI{XXXXXXX.XXXXXXX}
\acmConference[ICPP]{International Conference on Parallel Processing}{September 08--11,
	2025}{San Diego, CA}

\acmISBN{978-1-4503-XXXX-X/2018/06}

\begin{document}

	\title{Automatic Task Detection and Heterogeneous LLM Speculative Decoding}
	
	
\author{DanYing Ge}
\affiliation{%
	\institution{Beijing Normal University}
	\city{Beijing}
	\country{China}
}
\email{gdanying@mail.bnu.edu.cn}

\author{JianHua Gao}
\affiliation{%
	\institution{Beijing Normal University}
	\city{Beijing}
	\country{China}
}
\email{gaojh@bnu.edu.cn}

\author{QiZhi Jiang}
\affiliation{%
	\institution{Beijing Normal University}
	\city{Beijing}
	\country{China}
}
\email{202422081029@mail.bnu.edu.cn}

\author{YiFei Feng}
\affiliation{%
	\institution{Beijing Normal University}
	\city{Beijing}
	\country{China}
}
\email{202111030012@mail.bnu.edu.cn}

\author{WeiXing Ji}
\affiliation{%
	\institution{Beijing Normal University}
	\city{Beijing}
	\country{China}
}
\email{jwx@bnu.edu.cn}
	
	\begin{abstract}
		Speculative decoding, which combines a draft model with a target model, has emerged as an effective approach to accelerate large language model (LLM) inference. However, existing methods often face a trade-off between the acceptance rate and decoding speed in downstream tasks due to the limited capacity of the draft model, making it difficult to ensure efficiency across diverse tasks. To address this problem, we propose a speculative decoding algorithm tailored for downstream task optimization. It includes an automatic task partitioning and assigning method, which automatically categorizes downstream tasks into different sub-tasks and assigns them to a set of heterogeneous draft models. Each draft model is aligned with the target model using task-specific data, thereby enhancing the consistency of inference results. In addition, our proposed method incorporates an online lightweight prompt classifier to dynamically route prompts to the appropriate draft model. Experimental results demonstrate that the proposed method improves draft accuracy by 6\% to 50\% over vanilla speculative decoding, while achieving a speedup of 1.10$\times$ to 2.64$\times$ in LLM inference.
	\end{abstract}


	
	\maketitle
	
	\section{Introduction}
\begin{figure}[h]
		\centering
		\includegraphics[width=\linewidth]{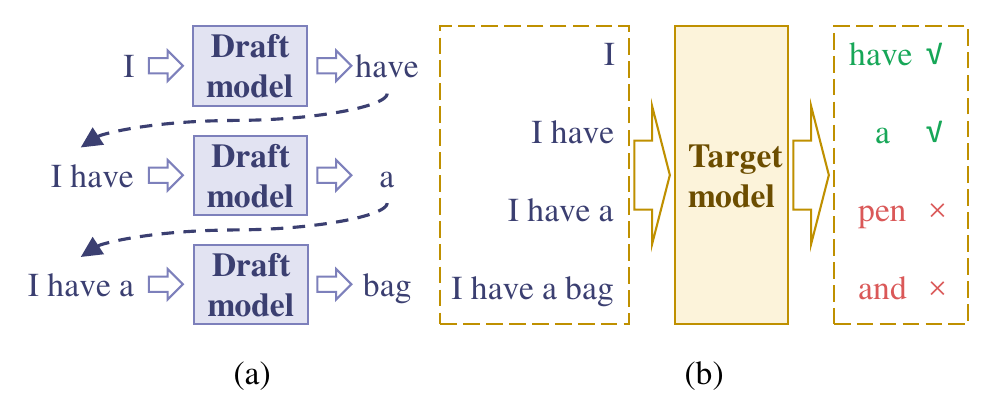}
		\caption{Workflow of speculative decoding. (a) Autoregressive decoding of the draft model. (b) Parallel verification of the target model.}
		\label{fig1}
		\Description{A woman and a girl in white dresses sit in an open car.}
	\end{figure}

	The scaling law shows that as model size, data, and computational resources increase, model performance improves in a predictable manner \cite{brown2020languagemodelsfewshotlearners, zhai2022scalingvisiontransformers, kaplan2020scalinglawsneurallanguage}. Mainstream large language models (LLMs) have parameter counts reaching tens of billions or even hundreds of billions, which leads to significant computational resource consumption during inference. Moreover, transformer-based LLMs typically employ an autoregressive decoding mechanism to sequentially generate tokens, in which the generation of the current token depends on all previous tokens. Speculative decoding, as presented in Figure~\ref{fig1}, was proposed by Google and DeepMind to improve the efficiency of LLM generation \cite{10.5555/3618408.3619203, chen2023acceleratinglargelanguagemodel}. It introduces a draft model that performs multiple consecutive autoregressive decoding steps to generate candidate tokens (Figure~\ref{fig1}-(a)), which are then validated in parallel by the target model (Figure~\ref{fig1}-(b)). Previous works show that speculative decoding has emerged as a new decoding paradigm for LLM inference \cite{xia2024unlockingefficiencylargelanguage}. 
	
	Although speculative decoding is promising in speeding up LLM inference, a careful balance must be struck between the drafting speed and acceptance rate of the draft model. The theoretical speedup of speculative decoding over autoregressive decoding is given by Equation (\ref{equ:speedup}) \cite{10.5555/3618408.3619203},
	\begin{equation}\label{equ:speedup}
		\frac{1 - \alpha^{\gamma + 1}}{(1 - \alpha)(\gamma c + 1)} \tag{1}
	\end{equation}
	where $\alpha$ is a measure of how well the draft model approximates the target model, and $c$ denotes the ratio of the time taken by the draft model to perform a single inference to the time taken by the target model to do the same work. It can be seen as an indicator of the inference cost of the draft model. Moreover, $\gamma$ denotes the number of tokens continuously generated by the draft model in one iteration. Given $\gamma$, the speedup is constrained by the drafting accuracy and the inference cost of the draft model. Higher drafting accuracy results in fewer verification failures from the target model, thus enhancing overall performance. However, according to the scaling law, achieving higher drafting accuracy typically requires larger or more complex draft models, which in turn increases the drafting cost.
	
	Current optimizations in speculative decoding primarily focus on optimizing either the drafting method or the parallel validation of the target model to improve performance of general tasks  \cite{Miao_2024, 10.5555/3692070.3692631, spector2023acceleratingllminferencestaged, 10.5555/3666122.3667436, 10.5555/3692070.3693328, chen2024cascadespeculativedraftingfaster, gong-etal-2024-graph, mamou2024dynamicspeculationlookaheadaccelerates, hooper2024speedspeculativepipelinedexecution, zhang-etal-2024-draft}. For example, Specinfer introduces collective boost-tuning and token tree validation to reduce end-to-end latency while maintaining text generation quality \cite{Miao_2024}. Online speculative decoding improves the accuracy of draft tokens by dynamically updating the draft model \cite{10.5555/3692070.3693328}. Other approaches improve speculative decoding by optimizing the drafting strategy, modifying the number of candidate tokens, implementing staged speculative decoding, and using the target model itself to generate draft tokens \cite{10.5555/3692070.3692631, chen2024cascadespeculativedraftingfaster, 10.5555/3666122.3667436, gong-etal-2024-graph,mamou2024dynamicspeculationlookaheadaccelerates, spector2023acceleratingllminferencestaged, hooper2024speedspeculativepipelinedexecution, zhang-etal-2024-draft}. For more details on speculative decoding optimization, please refer to the survey paper \cite{xia2024unlockingefficiencylargelanguage}.
	
	The performance of a language model is strongly correlated with its scale. On the one hand, a larger number of parameters tends to yield better performance on general tasks. On the other hand, an increase in model scale also leads to higher computational complexity, thereby increasing inference costs. Consequently, the drafting accuracy and inference efficiency of the draft model are inherently constrained by each other. Achieving a balance between drafting accuracy and inference efficiency for general tasks remains a significant challenge. Therefore, this work focuses on the domain-specific speculative decoding optimization.
	
\begin{figure}[h]
		\centering
		\includegraphics[width=\linewidth]{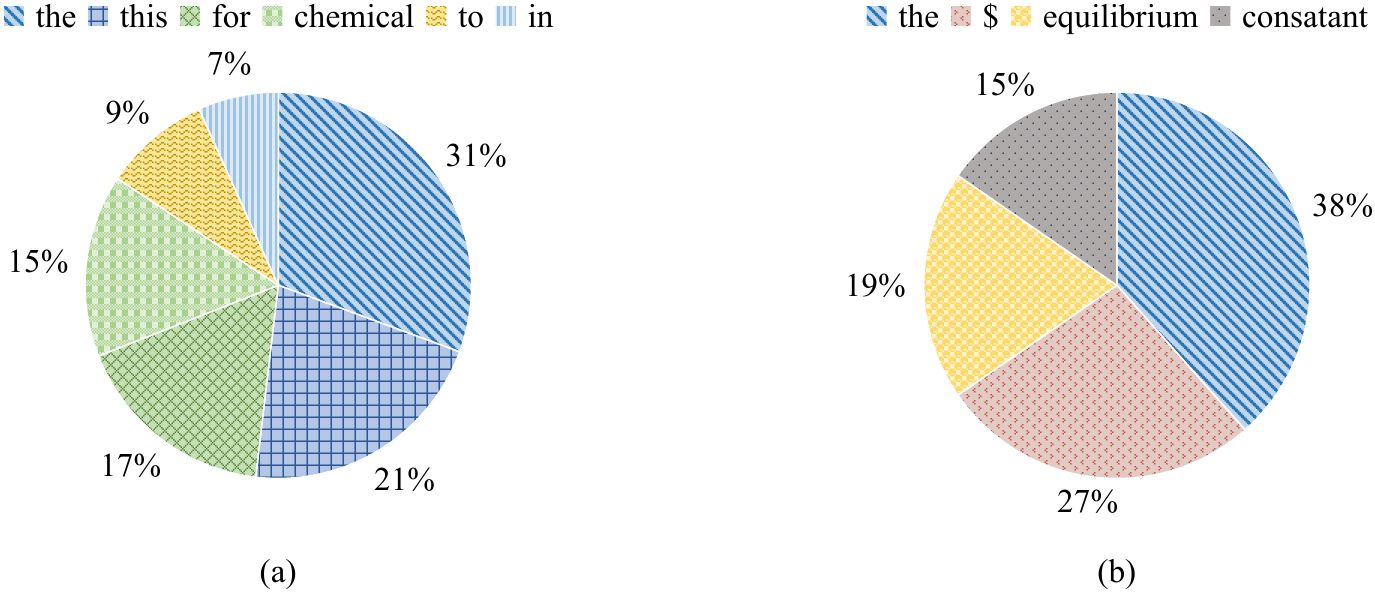}
		\caption{The word "reaction" tends to co-occur with different tokens in the domains of chemistry (a) and mathematics (b).}
		\label{fig-reaction}
	\end{figure}

	The association between a token and other tokens varies across different domains. In domain-specific contexts, the distribution patterns and co-occurrence relationships of tokens differ from those in general tasks, resulting in distinct semantic associations for the same token across domains. For example, a token that frequently co-occurs with certain terms in one domain may exhibit entirely different co-occurrence patterns in another domain. Take the mathematics domain as an example: the token "reaction" often appears alongside terms like "\$", whereas this association is rarely observed in chemistry-related contexts (Figure~\ref{fig-reaction}). This context-dependent semantic variability highlights the domain-specific nature of token representations and provides compelling motivation for conducting domain-specific research.
	
	In this paper, we propose TaskSpec, a speculative decoding approach tailored to downstream tasks in LLMs. By enhancing the acceptance rate of the draft model for downstream tasks, TaskSpec improves overall efficiency. TaskSpec is an automated speculative decoding system designed for downstream tasks, requiring no human intervention. First, inputs are processed by both the base draft model and the target model through vanilla speculative decoding to generate outputs. The input-output pairs are then used to construct a collected dataset and the system performs task-oriented clustering on this dataset, with each clustered dataset corresponding to a different task. Next, we customize draft models for different downstream tasks based on these clustered datasets. Finally, an online lightweight prompt classifier is employed to automatically select the most suitable draft model for various user inputs. By constructing a set of heterogeneous draft models, TaskSpec significantly improves the acceptance rate of the draft model in downstream tasks, while automating the entire process greatly enhances the method's transferability.
	
	The primary contributions of this paper can be summarized as follows:
	\begin{itemize}
		
		\item We propose a data-driven automated task partitioning method. In an inference system that runs continuously based on vanilla speculative decoding, we first collect user inputs and model outputs to construct a comprehensive dataset. We then employ clustering algorithms to automatically partition this dataset into multiple task-specific clusters, where each sub-dataset represents a distinct downstream task category. This approach enables fine-grained, domain-specific task partition while minimizing human intervention.
		\item We propose a heterogeneous draft models-based speculative decoding optimization method. Based on the clustered task-customized dataset, we trained a set of heterogeneous draft models, each specialized in handling a class of downstream tasks. In the process of online speculative decoding, we also designed a lightweight and accurate prompt classifier to dynamically select the most suitable model for each user request, improving the drafting accuracy of the draft model with minimal overhead.
		\item Experimental results based on LLaMA-2-13B and LLaMA-68M are presented, and the results demonstrate that the proposed TaskSpec achieves speedups of up to 1.5$\times$, 2.6$\times$, 1.8$\times$, and 1.8$\times$ over the vanilla speculative decoding on four tasks, respectively.
	\end{itemize}
	
	The remainder of the paper is organized as follows. Section 2 introduces the TaskSpec proposed in this paper. Section 3 presents comprehensive experimental evaluations and in-depth discussions. Section 4 covers related work on speculative decoding, and Section 5 concludes this work.
	
	\section{TaskSpec}
	\subsection{Overview}
	
\begin{figure}[ht]
		\vskip -0.1in
		\begin{center}
			\centerline{\includegraphics[width=\columnwidth]{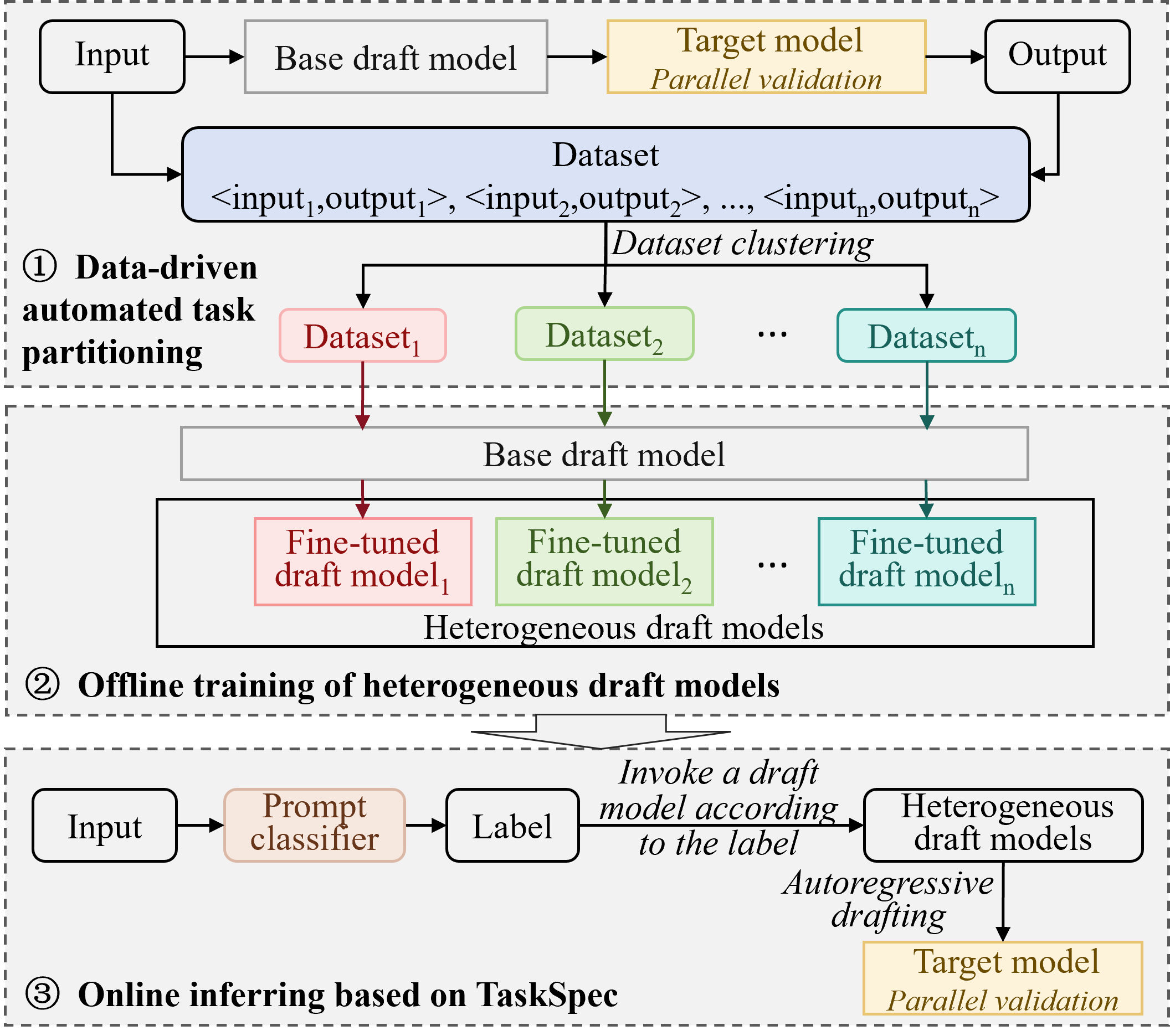}}
			\caption{The overview of TaskSpec.}
			\label{fig2}
		\end{center}
		\vskip -0.3in
	\end{figure}

	Draft models with limited parameters struggle to achieve performance similar to target models in general tasks. However, increasing the parameter of the draft model also raises the computational cost of autoregressive decoding, which offsets the performance of speculative decoding.  
	To address the challenge of balancing accuracy and efficiency of draft models, we propose TaskSpec, a downstream task-oriented LLM speculative decoding approach. The overview of TaskSpec is presented in Figure~\ref{fig2}.
	
	Firstly, TaskSpec performs vanilla speculative decoding on the user inputs. After a period of continuous collection of user inputs and speculative decoding outputs, we construct a collected dataset. We then conduct clustering analysis on this dataset, partitioning it into multiple clustered datasets, each corresponding to a downstream task with similar semantic features or application scenarios. Based on these clustered datasets, we fine-tune the base draft model and construct a set of heterogeneous draft models tailored to different tasks. During inference, if the user input specifies a task type, it is directly routed to the corresponding draft model for processing. If the task type is not specified, a lightweight online prompt classifier is used to identify the task and route the input to the most suitable draft model for subsequent speculative decoding.
	
	The entire process of TaskSpec exhibits a high level of automation, requiring minimal human intervention, and can easily be applied to different domains and tasks. Meanwhile, the combination of the heterogeneous draft model set and the lightweight online prompt classifier significantly improves the acceptance rate of the draft models, thereby further enhancing decoding efficiency.
	
	\subsection{Data-driven Automated Partitioning of Downstream Tasks}
	
\begin{figure}[ht]
		\vskip -0.05in
		\begin{center}
			\centerline{\includegraphics[width=0.85\columnwidth]{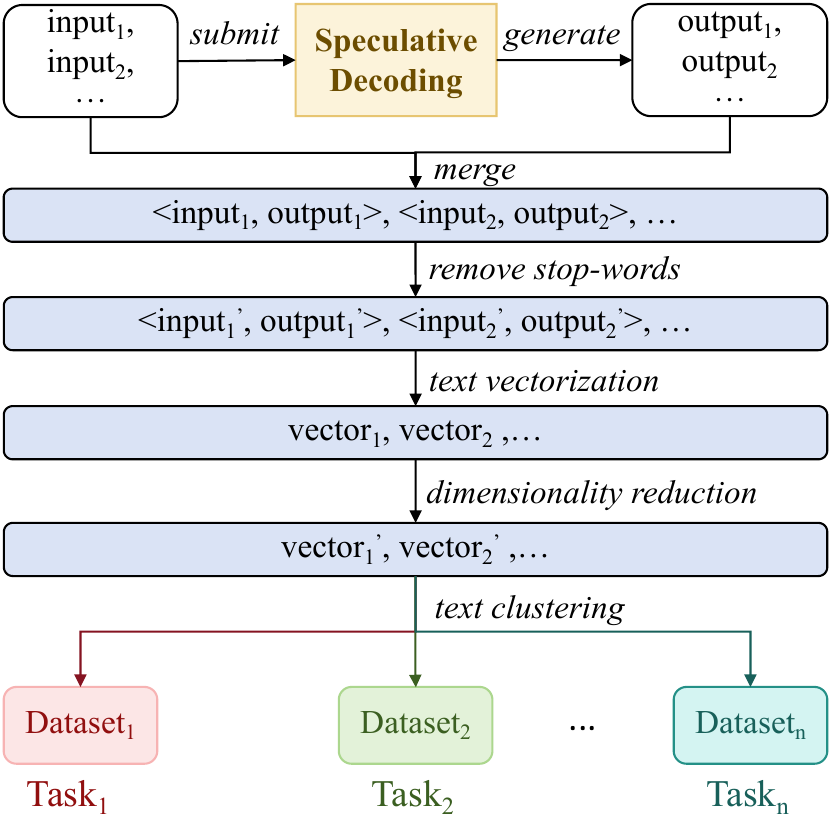}}
			\caption{Data-driven automated partitioning of downstream tasks.} 
			\label{fig3}
		\end{center}
	\end{figure}

	To enhance the transferability of this method across different domains, we propose to automatically build task-specific datasets for draft model fine-tuning. Figure~\ref{fig3} illustrates the overall process.
	
	For each user input, the corresponding output can be obtained through vanilla speculative decoding, forming a collected dataset composed of \(\textless {input,output}\textgreater{ }\) pairs. Next, we partition this collected dataset into multiple clustered datasets, each oriented toward a different downstream task. Specifically, we first remove stop words and special characters to enhance the quality of text representation and reduce irrelevant noise. Then, we represent the text as vectors and reduce their dimensionality to lower complexity while removing redundant information. Subsequently, we apply the clustering algorithm to the preprocessed dataset, partitioning it into multiple subsets denoted as  \( D = \{ d_1, d_2, d_3, \dots \}\), which correspond to the downstream task set \( T = \{ t_1, t_2, t_3, \dots\} \). The samples within each subset exhibit high semantic consistency, facilitating the training of more task-specific draft models. This, in turn, enhances the accuracy of draft generation during inference and improves the overall efficiency of speculative decoding.
	
	\subsection{Construction of Heterogeneous Draft Models}
	A draft model with limited parameters is difficult to adapt to multiple tasks simultaneously and achieve performance consistent with the target model across all tasks. Therefore, we propose building heterogeneous draft models tailored to different downstream tasks. It aligns the inference performance of draft models with that of the target model for each specific downstream task, enabling different draft models to achieve high acceptance rate on corresponding tasks.
	
\begin{figure}[ht]
		\vskip -0.1in
		\begin{center}
			\centerline{\includegraphics[width=0.85\columnwidth]{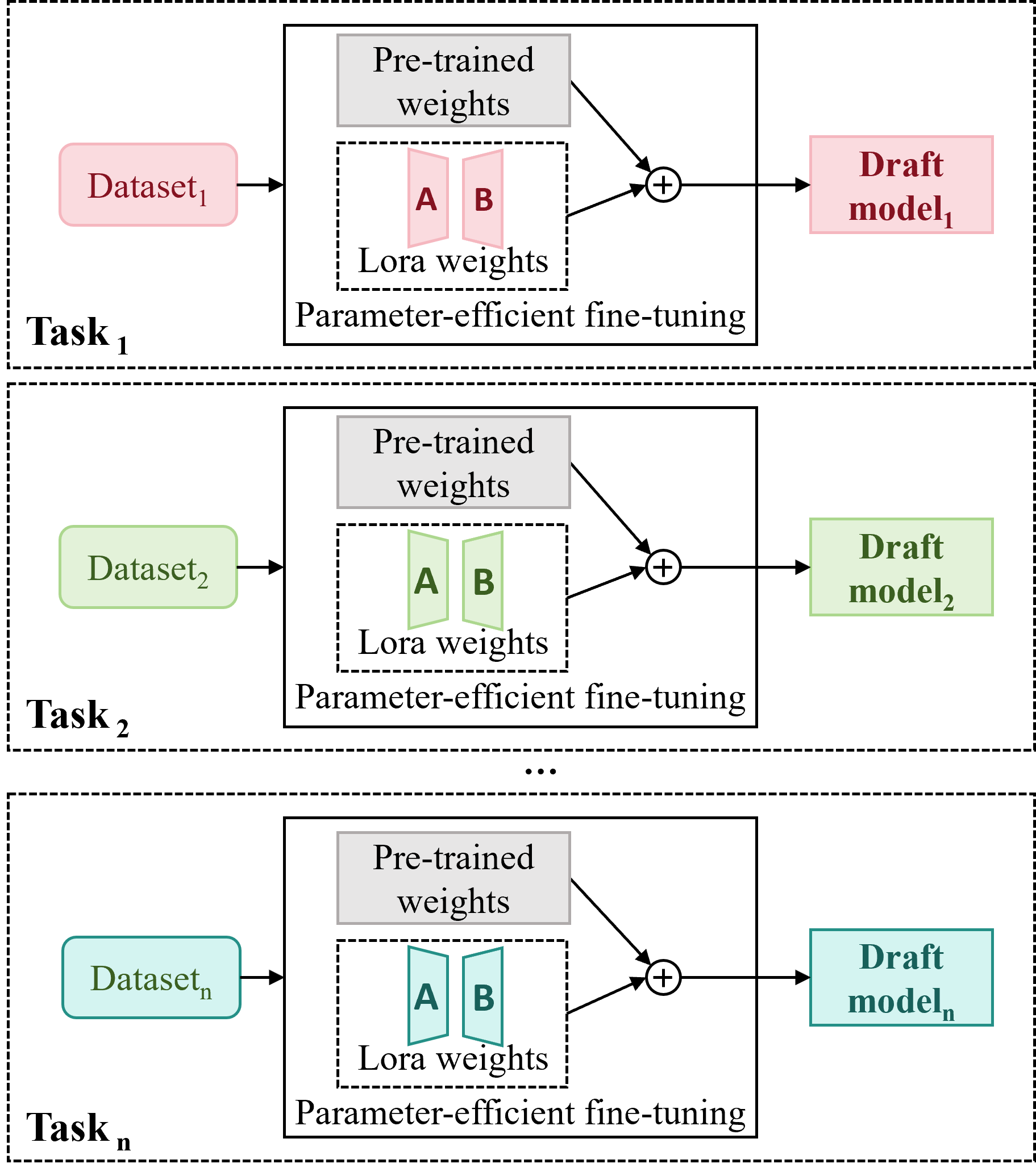}}
			\caption{Construction of heterogeneous draft models for downstream tasks.}
			\label{fig4}
		\end{center}
		\vskip -0.3in
	\end{figure}

	Figure \ref{fig4} illustrates the overall process of constructing a set of heterogeneous draft models for downstream tasks. 
	Based on the same base draft model, we train task-specific draft models for different downstream tasks, aiming to minimize the output discrepancies between the draft model and the target model on the same task. Each draft model is fine-tuned using the Low-Rank Adaptation (LoRA) \cite{hu2021loralowrankadaptationlarge}. 
	
	In summary, for each type of task, the primary goal of the above supervised fine-tuning (SFT) is to align the inference behavior of the draft model as closely as possible with that of the target model, ensuring that each draft model can achieve high guessing accuracy for the corresponding task. By fine-tuning the base draft model for each task individually, a set of heterogeneous draft models tailored for different downstream tasks can be constructed, with each model demonstrating strong performance for its respective task. Finally, in multi-task scenarios, the fine-tuned draft models can generate higher-quality tokens that are more similar to the outputs of the target model for the respective tasks, achieving higher guessing accuracy and thereby improving decoding efficiency.
	
	\subsection{Online Lightweight Prompt Classifier}\label{sec:classifier}
	In multi-task scenarios, the specific task category of an input prompt may not necessarily be known, making it challenging to figure out which draft model should be used in conjunction with the target model. To address this issue, we design an online, lightweight prompt classifier that classifies input prompts based on their features. The classifier generates the task category of the input prompt and then calls the corresponding draft model for speculative decoding. It enables us to call the most suitable draft model for speculative decoding to each user prompt, thereby improving the overall efficiency.
	
	The prompt classifier proposed in this paper requires two steps: offline training and online decision-making. In the offline training phase, we first preprocess the input prompts by performing tokenization and removing stop words to make the task-related features more distinct and easier to capture. Then, based on the preprocessed dataset, we train a text classification model. The architecture of the model comprises an embedding layer, a Mamba-based sequence encoding layer, a pooling layer, and a classification layer \cite{mamba, mamba2}. Notably, we innovatively adopt Mamba as the core sequence modeling module instead of multilayer perceptron (MLP) or Transformer blocks. Addressing the limited receptive field of traditional MLP and the quadratic computational complexity inherent in Transformers, Mamba achieves significantly accelerated inference speeds with linear scaling in sequence length, while maintaining competitive performance on real-world data up to million-token-scale sequences. This hybrid architecture preserves linear computational complexity through parameterized projection networks that enable global dynamic context modeling. Compared with fixed-weight MLP structures, our solution exhibits enhanced capability in capturing long-range semantic dependencies, thereby providing a more efficient framework for long-text classification tasks. We retrain it to adapt to the classification of specific downstream tasks. In the online decision-making phase, we first preprocess the user’s input prompt to reduce noise and enhance the classification model's focus on the core semantic information. Afterward, the text is tokenized and fed into the retrained model for prediction, enabling efficient and accurate prompt classification.
	
\begin{figure}[ht]
		\vskip -0.1in
		\begin{center}
			\centerline{\includegraphics[width=\columnwidth]{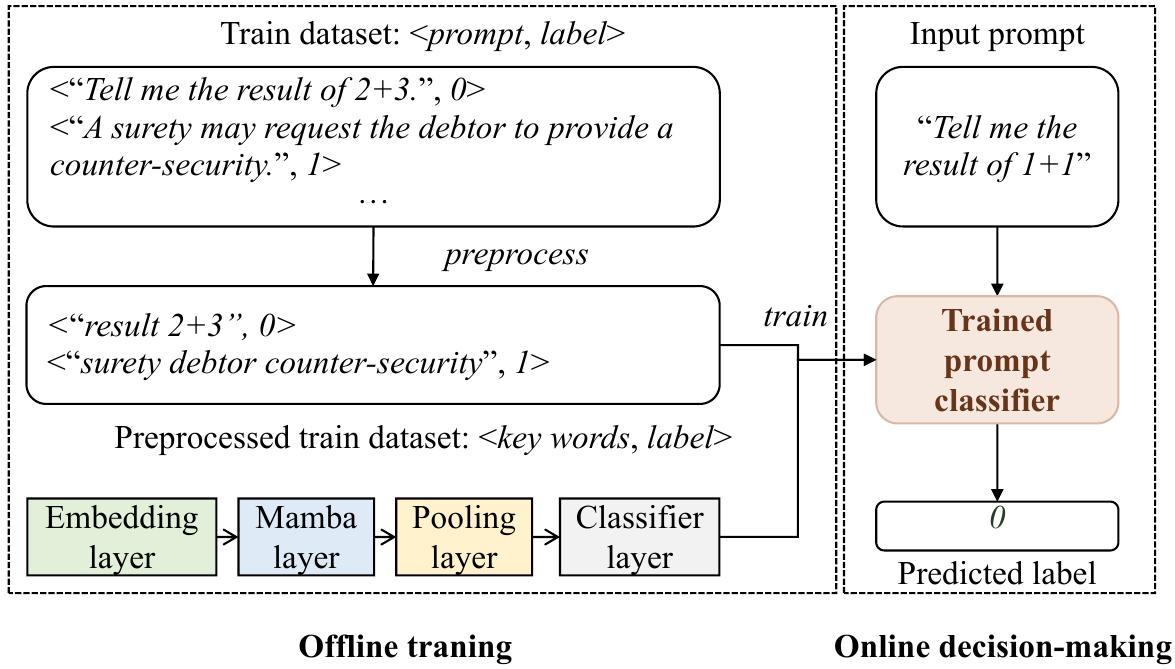}}
			\caption{Offline construction and online decision-making of the lightweight prompt classifier.}
			\label{fig5}
		\end{center}
		\vskip -0.3in
	\end{figure}

	Figure~\ref{fig5} illustrates an example of the prompt classification process. In the offline training phase, the task category of the first prompt “\textit{Tell me the result of 2+3}” is 0, and the task category of the second prompt “\textit{A surety may request the debtor to provide a counter-security}” is 1. After preprocessing the first prompt, we obtain the keywords “\textit{result}” and “\textit{2+3}”. The preprocessing results of the second prompt are “\textit{surety}”, “\textit{debtor}”, and “\textit{counter-security}”. Using the preprocessed results and their corresponding labels, we retrain the original classification model to obtain a prompt classifier for the multi-task scenarios. In the online decision-making phase, supposing the input is “\textit{Tell me the result of 1+1}”, the prompt classifier classifies it with a result of label 0.
	
	Compared with the Mixture of Experts (MoE) structure, where the expert models are fixed and lack flexibility in adapting to task variations, TaskSpec introduces task-specific data clustering, enabling dynamic adjustment of the draft model set based on the number and type of tasks, while ensuring complete independence between the draft models \cite{jacobs1991adaptive}. This mechanism allows TaskSpec to achieve higher adaptability and precision across different tasks. In terms of expert selection, MoE typically relies on a gating network for static or dynamic routing of experts without distinguishing between tasks, whereas TaskSpec innovatively uses a lightweight online prompt classifier to dynamically select draft models based on task features. Overall, compared with MoE, TaskSpec emphasizes enhancing the efficiency and flexibility of the inference stage through task-specific fine-tuning and optimized inference, thus achieving a more refined balance between inference speed and accuracy.
	
	Beyond the comparison with MoE architectures, TaskSpec also differs substantially from existing optimization methods in its practical design philosophy. Unlike existing studies, the optimization method proposed in this paper is specifically tailored for downstream tasks, balancing the acceptance rate and inference efficiency of the draft model across different tasks. TaskSpec not only implements dynamic task partitioning with strong cross-domain transferability but also constructs a heterogeneous set of draft models. By fine-tuning the draft models for specific downstream tasks, it significantly improves the draft stage's prediction acceptance rate without increasing inference overhead.
	
	In summary, to improve the quality of draft model generation while aligning its performance with the target model as much as possible, we propose a downstream task-oriented speculative decoding optimization method. It helps improve the overall efficiency of speculative decoding across various domains. Our approach can dynamically determine the number of tasks and construct LLM-aligned datasets for different tasks and trains heterogeneous draft models based on these datasets to improve the accuracy of speculative predictions for different tasks. Considering that the task category of the input prompt may be unknown in online multi-task scenarios, a lightweight prompt classifier is designed to assist in online decision-making. By selecting the most suitable draft model for each task, TaskSpec can achieve higher speculative acceptance rate, thereby improving the overall inference efficiency of the system.
	
	\section{Experiments}
	\subsection{Experimental Setup}
	
	\textbf{Software and Hardware:} All experiments were performed on a server equipped with two Intel Xeon Platinum 8336C CPUs and one NVIDIA A100 GPU. The server has 516 GB of host memory and 40 GB of GPU memory. It runs the operating system of Ubuntu 22.04, the deep learning framework of PyTorch 2.0.1, and CUDA Toolkit 11.5.
	
	\noindent\textbf{Models and Datasets:} The target model used in the experiments is LLaMA-2-13B, with LLaMA-68M serving as the base draft model \cite{touvron2023llama2openfoundation, Miao_2024}. In the subsequent experiments, we take representative tasks from the education domain as examples to verify the effectiveness of TaskSpec. These tasks include text generation (Chinese), logical reasoning (Math), translation (English), and question answering (Chemistry). 
	
	We use data in Wanjuan 1.0 and ChemData700K to simulate the input data stream in online inference \cite{he2023wanjuancomprehensivemultimodaldataset, he2024opendatalabempoweringgeneralartificial, zhang2024chemllm}. Wanjuan 1.0 is a high-quality open-source multimodal dataset of over 2 TB jointly released by OpenDataLab. It encompasses various modalities, such as text, image-text, and video, and covers a wide range of domains, including science, literature, and education. The dataset offers high reliability and broad adaptability, making it a valuable resource for pretraining large models \cite{chen2024far, gu2024mllmguard}. We extract 95,659, 159,445, and 219,546 prompts from grades 7, 8, and 9 in the \textit{exam} subset to build Chinese, Math, and English-specific datasets. ChemData700K is an instruction fine-tuning dataset designed to enhance large language models’ chemistry capabilities.
	
	\noindent\textbf{Comparison Methods:} We compare our proposed method with three other approaches: autoregressive decoding, original speculative decoding, and REST \cite{10.5555/3618408.3619203, he2023rest}. Autoregressive decoding is the most fundamental decoding method, which generates each token sequentially until a termination token is encountered, serving as the performance baseline in our experiments. The original speculative decoding uses a small-scale draft model to draft multiple tokens sequentially, which are then validated in parallel by the target model \cite{10.5555/3618408.3619203}, which is referred to as "Vanilla SpecDec" in the following experiments. REST is also a domain-oriented speculative decoding method that generates draft tokens by retrieving a domain-specific knowledge base.
	
	\noindent\textbf{Evaluation Metrics:} We use the acceptance rate of the draft model, the total inference latency, and average acceptance length as metrics to evaluate different approaches. For REST, since the generation of draft tokens does not require inference from a draft model, we only include it in the comparison of average Acceptance Length, where computational efficiency is not directly impacted by model inference.
	\begin{itemize}
		\item Acceptance rate: This metric is used to measure the consistency between the texts generated by the draft model and the target model, which is given by Equation~(\ref{eq2}).
		\vskip -0.1in
		\begin{equation}
			\text{Acceptance Rate}=\frac{\text{Total number of accepted draft tokens}}{\text{Total number of  draft tokens}} \tag{2}
			\label{eq2}
		\end{equation}
		\item Walltime speedup: The actual test speedup relative to Autoregressive decoding.
		\item Average acceptance length $\tau$: The average number of tokens accepted per forward pass of the target model.
	\end{itemize}

	\noindent\textbf{Parameter Settings:} All experiments are conducted with a batch size of 1 and use the Greedy Sampling. The target model validation uses the non-deterministic verification proposed by Google \cite{10.5555/3618408.3619203}. Furthermore, to ensure the reliability of the experimental results, all experiments are independently repeated three times on the same experimental platform, with the average as the final reported results.
	
	\subsection{Main Results}
	To determine the size of the fine-tuning dataset used in our experiments, we evaluate its impact on the inference efficiency and the number of fine-tuning prompt-response pairs is set to 8,192 according to our evaluation. Besides, we analyzed the length of responses in various tasks and set the length to 128 for consistency across tasks.
	
	\begin{figure}
		\begin{center}
			\centerline{\includegraphics[width=\columnwidth]{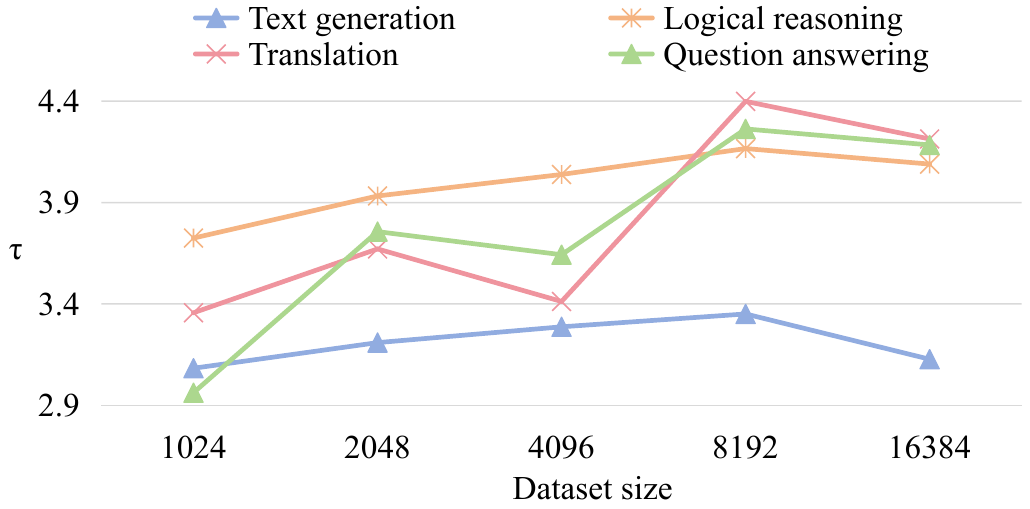}}
			\caption{ The impact of fine-tuning dataset size on the average acceptance length of speculative decoding across text generation, logical reasoning, translation, and question answering tasks.}
			\label{fig9}
		\end{center}
		\vskip -0.2in
	\end{figure}
	
	To evaluate the impact of fine-tuning dataset size on the inference efficiency of speculative decoding, we collected the average length of accepted tokens per iteration of TaskSpec on the test set using 1,024, 4,096, 8,192, and 16,384 prompt-response pairs to fine-tune draft models for different tasks, as shown in Figure \ref{fig9}-(a). We can observe that as the fine-tuning dataset size increases from 1,024 to 4,096, the average length gradually inclines with the relatively slow trend. When the dataset size is further increased to 8,192, the average length on text generation and logical reasoning inclines significantly. However, upon increasing the dataset size to 16,384, due to overfitting-prone, the average length on text generation and logical reasoning tasks begins to decrease, while that for translation and question answering continues to incline \cite{wang2024lorameetsdropoutunified}. This indicates that fine-tuning with a dataset of appropriate size can improve the accuracy of the draft model, thereby improving the overall efficiency of speculative decoding. Beyond a certain threshold, however, limitations such as overfitting or the fine-tuning method may cause the model's acceptance rate to decrease, leading to an increase in the overall inference cost. In conclusion, the fine-tuning dataset size is uniformly set to 8,192.

	In the subsequent experiments, we fine-tuned the base draft model using the LLM-aligned datasets for different tasks, resulting in draft models tailored to handle  text generation, logical reasoning, translation and question answering tasks. We then tested performance on the test sets using the online classifier and heterogeneous draft models, recording the acceptance rate of the draft models and the total inference latency with different $\gamma$. Here, $\gamma$ represents the number of continuous autoregressive decoded tokens by draft models. We also tested the inference time of TaskSpec and original speculative decoding, as well as the acceptance rate of the draft model, at the $\gamma$  that achieved the best performance for each type of tasks.
	\subsubsection{Acceptance Rate of Draft Tokens}
	
\begin{figure}
		\begin{center}
			\centerline{\includegraphics[width=\columnwidth]{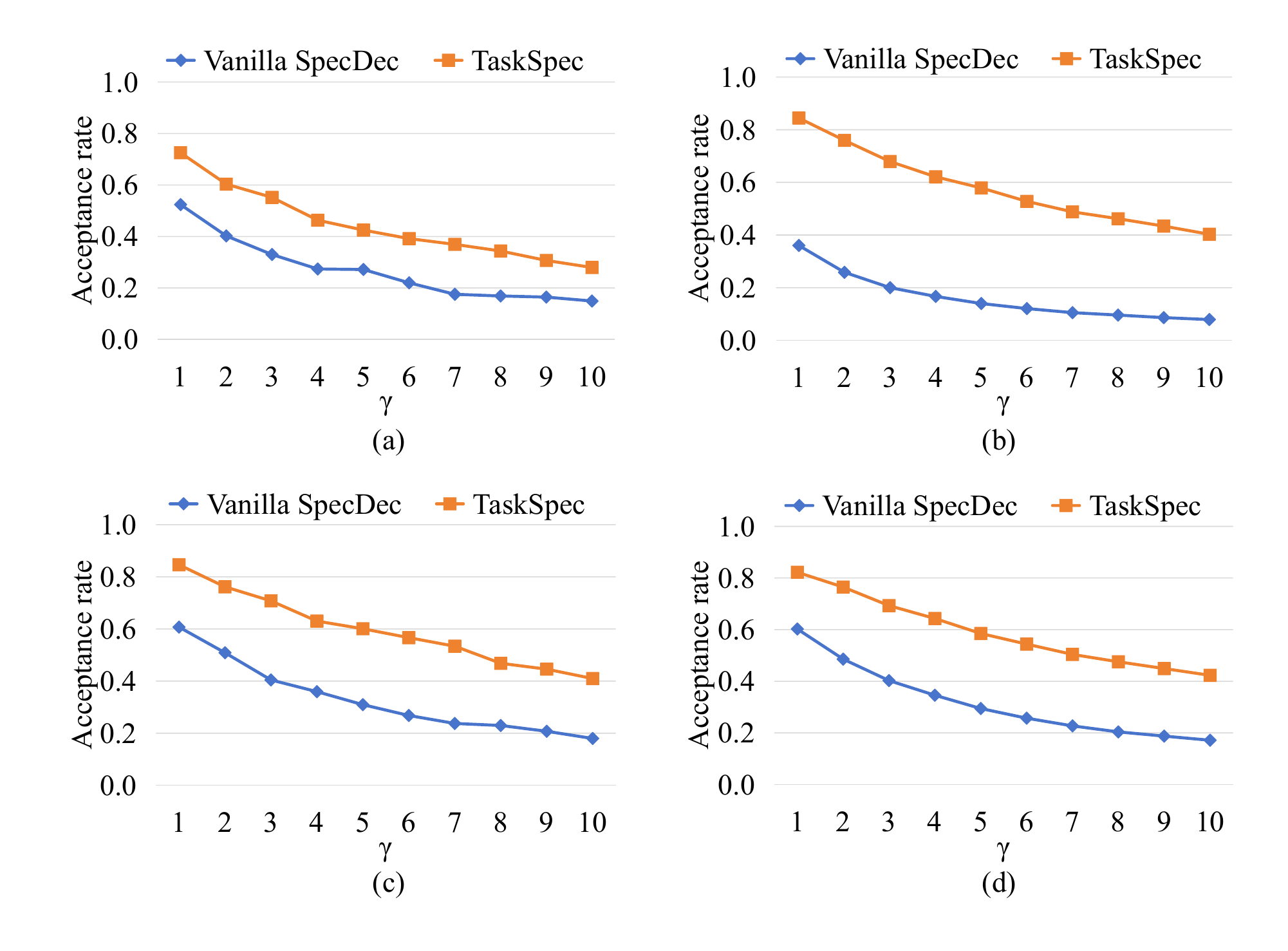}}
			\caption{Acceptance rate of Vanilla SpecDec and TaskSpec on (a) text generation, (b) logical reasoning, (c) translation and (d) question answering.}
			\label{fig6}
		\end{center}
		\vskip -0.2in
	\end{figure}

	Figure~\ref{fig6} presents the acceptance rate of the draft models in different speculative decoding methods on text generation, logical reasoning, translation and question answering. It can be observed that, TaskSpec significantly improves the acceptance rate of the draft models across different tasks. Notably, on logical reasoning task, the average acceptance rate of the vanilla speculative decoding is below 20\% when $\gamma$ is 4, whereas the draft model of TaskSpec achieves an average acceptance rate of over 60\%.
	Overall, for text generation, vanilla speculative decoding achieves an average of 27\% and a maximum of 52\% acceptance rate, while TaskSpec achieves an average of 45\% and a maximum of 73\%. For logical reasoning tasks, TaskSpec improves the average and maximum accuracy from 16\% and 36\% to 58\% and 84\%, respectively. On translation task, the average and maximum acceptance rate of the draft model in Vanilla SpecDec are 33\% and 61\%, respectively, while in TaskSpec, the average and maximum acceptance rate are 60\% and 85\%, respectively. On question answering task, the average and maximum acceptance rate of the draft model in Vanilla SpecDec are 32\% and 60\%, respectively,60\% and 82\% in TaskSpec, respectively.
	We can also observe that the accuracy of draft models gradually decreases as $\gamma$ increases in speculative decoding. The reason is that the acceptance probability of different tokens decreases as the distance from the first token increases \cite{chen2024cascadespeculativedraftingfaster}.
	
	\begin{figure}
		\begin{center}
			\centerline{\includegraphics[width=\columnwidth]{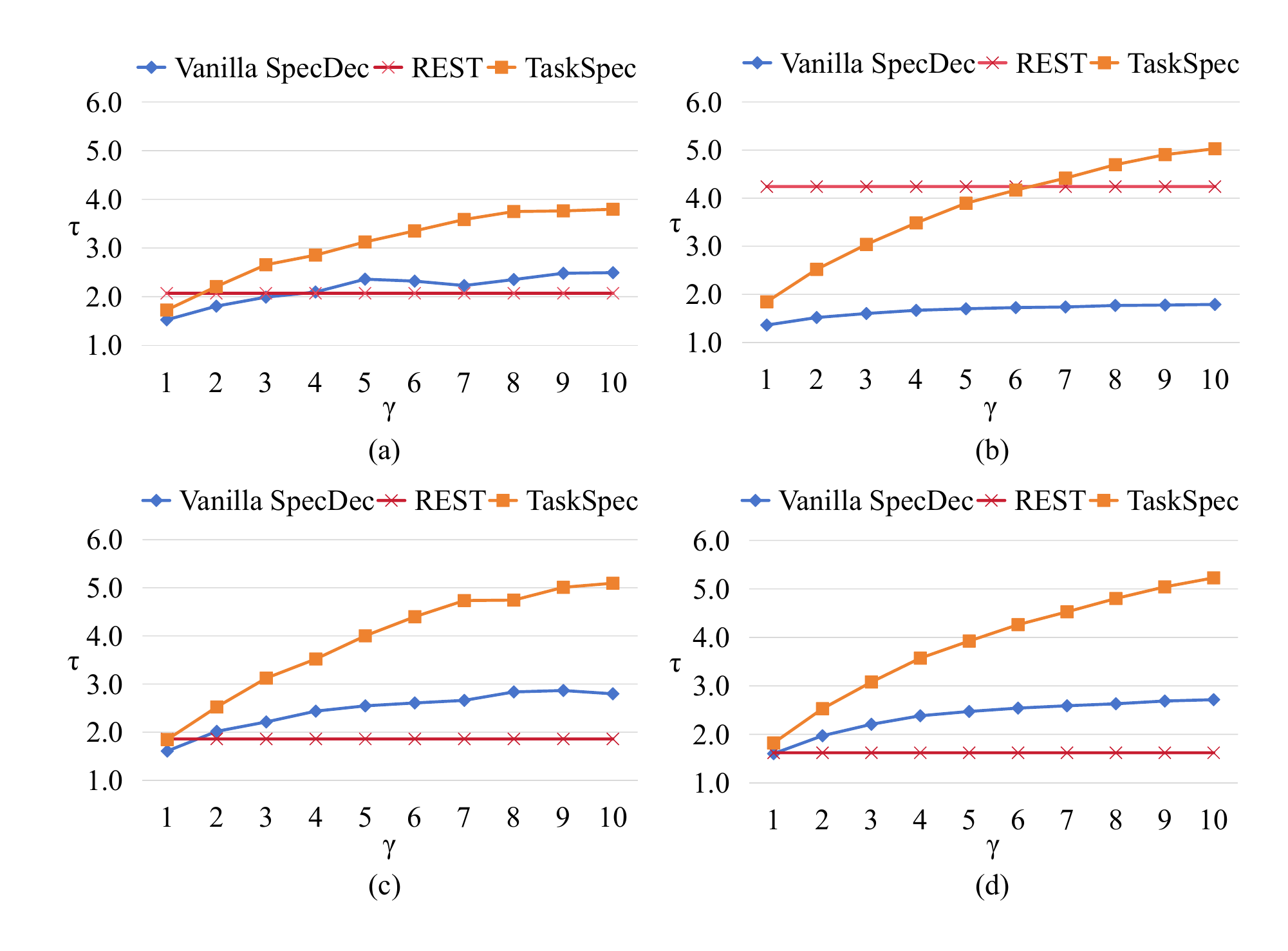}}
			\caption{Average acceptance length of Vanilla SpecDec, REST and TaskSpec on (a) text generation, (b) logical reasoning, (c) translation and (d) question answering.}
			\label{fig8}
		\end{center}
		\vskip -0.2in
	\end{figure}
	
	\subsubsection{Inference Overhead and Average Length of Accepted Tokens per Iteration}
	
\begin{figure}
		\begin{center}
			\centerline{\includegraphics[width=\columnwidth]{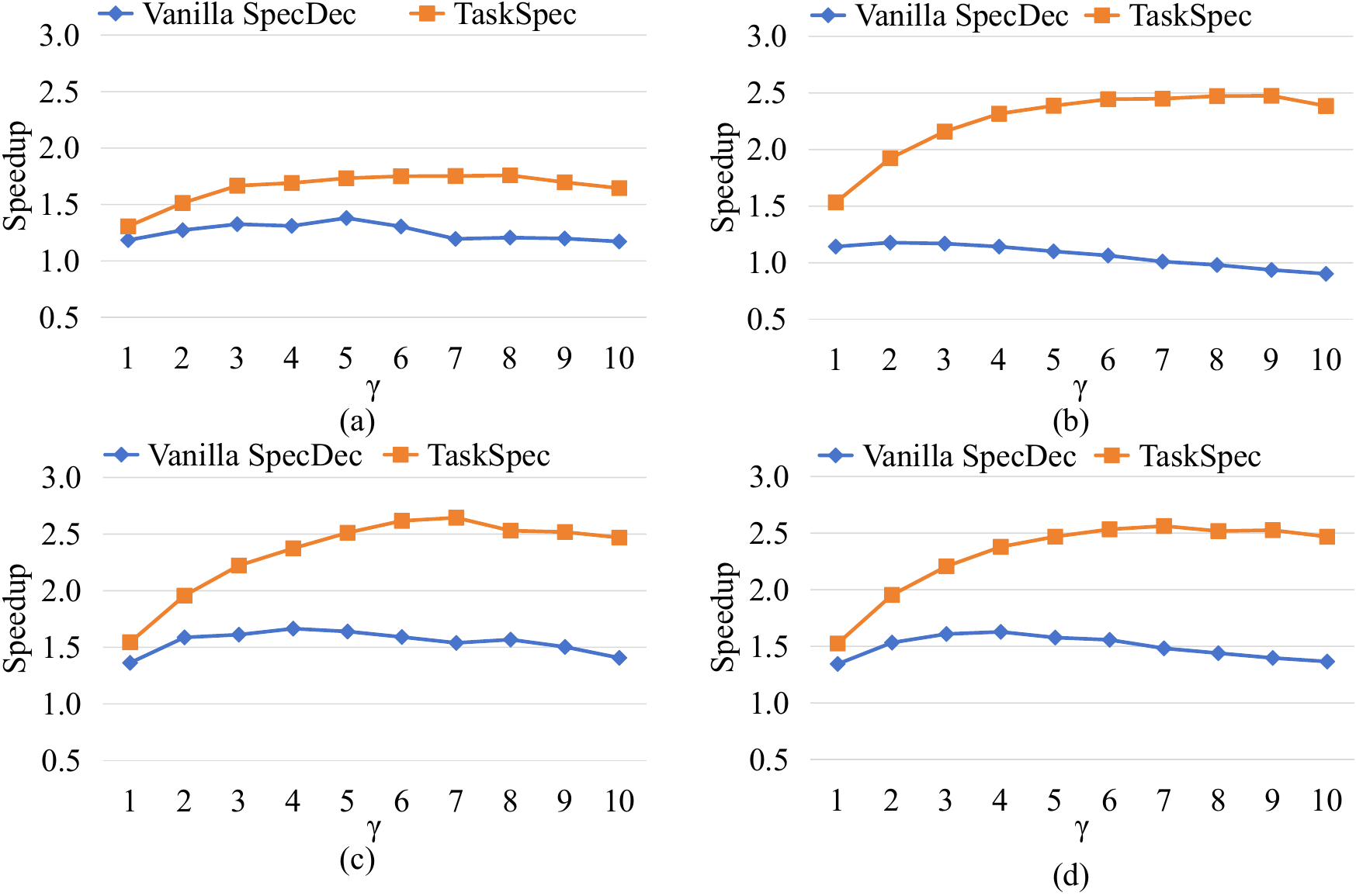}}
			\caption{Speedups of Vanilla SpecDec and TaskSpec over the LLM autoregressive decoding on (a) text generation, (b) logical reasoning, (c) translation and (d) question answering.}
			\label{fig7}
		\end{center}
		\vskip -0.2in
	\end{figure}

	Figure~\ref{fig8} and Figure~\ref{fig7} present the average length of accepted tokens per iteration and speeddups of the Vanilla SpecDec, REST and TaskSpec over the original autoregressive decoding on text generation(a), logical reasoning(b), translation(c) and question answering(d), as $\gamma$ varies. The experimental results indicate that TaskSpec reduces the inference time across different tasks. Compared with vanilla speculative decoding, TaskSpec achieves the highest speedups of 1.47$\times$, 2.64$\times$, 1.76$\times$, and 1.81$\times$ when $\gamma$ is set to 7, 10, 10, and 9, respectively, across four tasks. Additionally, TaskSpec improves the average accepted draft token length by 1.61$\times$, 2.81$\times$, 1.82$\times$, and 1.26$\times$ across the four tasks, respectively.
	
	Compared with REST, TaskSpec achieves up to a 3.23$\times$ improvement in the average accepted draft token length. This result indicates that TaskSpec significantly enhances the quality of the draft model's outputs, thereby reducing the verification overhead of the target model and further improving inference efficiency.
	
	By examining Figure~\ref{fig6} and Figure~\ref{fig7}, it can be observed that as $\gamma$ increases, the acceptance rate of draft tokens decreases, while the speedup exhibits an initial increase followed by a decline. This is attributed to the following reasons: When $\gamma$ is small, although the accuracy is high, the number of triggered target model verifications is also large. As $\gamma$ increases, the number of target model verifications decreases, leading to an acceleration in inference speed. However, when $\gamma$ exceeds a certain threshold, further increase in $\gamma$ leads to lower acceptance rate, and only a small fraction of the tokens generated by draft models are accepted by the target model. 
	
	\subsection{Ablation Studies}
	
	\subsubsection{Data-driven Automatic Partitioning of Downstream Tasks}
	
	We develop a dynamic task partitioning mechanism. Specifically, We perform data cleaning and preprocessing on the raw data, including the removal of special characters and stop-words. Then, we utilize paraphrase-multilingual-MiniLM-L12-v2 which maps sentences to a 384 dimensional dense vector space to encode the text into vector representations \cite{reimers-2019-sentence-bert}. Following preprocessing, we perform dimensionality reduction and apply K-means clustering algorithm to automatically partition the data in the feature space, thereby generating task-specific subsets. The experimental results show that when the number of tasks is 2, 3, and 4, our method achieves clustering accuracies of 99\%, 98\%, and 90.25\%, respectively, validating the effectiveness and reliability of the clustering-based task partitioning strategy.
	
	
	\begin{table*} 
		\caption{Ablation study results.}
		\label{ablation}
		\begin{tabularx}{\textwidth}{
				l l
				>{\centering\arraybackslash}X
				>{\centering\arraybackslash}X 
				>{\centering\arraybackslash}X 
				>{\centering\arraybackslash}X 
				>{\centering\arraybackslash}X }
			\toprule
			\textbf{$\gamma$}&\textbf{Metric} & \textbf{Method} & \textbf{Text generation} & \textbf{Logical reasoning} & \textbf{Translation} & \textbf{Question answering} \\
			\midrule
			\multirow{6}[0]{*}{9} & \multirow{2}[0]{*}{Speedup} & Unary & 1.61  & \multicolumn{1}{c}{2.05 } & \multicolumn{1}{c}{2.15 } & \multicolumn{1}{c}{2.05 } \\
			&       & TaskSpec & \underline{1.70}  & \multicolumn{1}{c}{\underline{2.48} } & \multicolumn{1}{c}{\underline{2.52} } & \multicolumn{1}{c}{\underline{2.53} } \\
			& \multirow{2}[0]{*}{Acceptance rate} & Unary & \multicolumn{1}{c}{0.28 } & \multicolumn{1}{c}{0.34 } & \multicolumn{1}{c}{0.36 } & \multicolumn{1}{c}{0.33 } \\
			&       & TaskSpec & \multicolumn{1}{c}{\underline{0.31} } & \multicolumn{1}{c}{\underline{0.43} } & \multicolumn{1}{c}{\underline{0.45} } & \multicolumn{1}{c}{\underline{0.45} } \\
			& \multicolumn{1}{c}{\multirow{2}[0]{*}{Average acceptance length}} & Unary & \multicolumn{1}{c}{3.53 } & \multicolumn{1}{c}{4.06 } & \multicolumn{1}{c}{4.20 } & \multicolumn{1}{c}{3.99 } \\
			&       & TaskSpec & \multicolumn{1}{c}{\underline{3.76} } & \multicolumn{1}{c}{\underline{4.90} } & \multicolumn{1}{c}{\underline{5.01} } & \multicolumn{1}{c}{\underline{5.04} } \\
			\midrule
			\multirow{6}[0]{*}{10} & \multirow{2}[0]{*}{Speedup} & Unary & 1.55  & 2.02  & 2.13  & 2.05  \\
			&       & TaskSpec & \underline{1.64}  & \underline{2.39}  & \underline{2.47}  & \underline{2.47}  \\
			& \multirow{2}[0]{*}{Acceptance rate} & Unary & 0.25  & 0.32  & 0.34  & 0.31  \\
			&       & TaskSpec & \underline{0.28}  & \underline{0.40}  & \underline{0.41}  & \underline{0.42}  \\
			& \multicolumn{1}{c}{\multirow{2}[0]{*}{Average acceptance length}} & Unary & 3.49  & 4.17  & 4.41  & 4.15  \\
			&       & TaskSpec & \underline{3.80}  & \underline{5.03}  & \underline{5.10}  & \underline{5.23}  \\
			
			\bottomrule
		\end{tabularx}
	\end{table*}
	
	\subsubsection{Heterogeneous Draft Model Sets}
	
	To evaluate the contribution of heterogeneous draft models, the following experiment utilized a comprehensive LLM-aligned dataset, which includes a total of 8192 prompt-response pairs related to the four tasks mentioned above. Using LLaMA-68M as the base draft model, we fine-tuned a unary draft model on this dataset. The same test set was then used to evaluate the performance of the unary draft model (labeled with "Unary") and the proposed multiple heterogeneous draft models (labeled with "TaskSpec"). Table~\ref{ablation} shows the acceptance rate, speedups, average length of accepted tokens for both approaches. It can be observed that without considering task-specific features, Unary performs relatively worse than our proposed method. TaskSpec achieves up to 12\% improvement in acceptance rate, 1.21$\times$ speedup, and an increase of 1.23$\times$ in the average number of accepted tokens per iteration compared with the Unary baseline.
	
	
	\subsubsection{Prompt Classifier}
	
	We trained a prompt classifier using 1,026,401 prompt-label pairs and tested it on a dataset of size 400. As described in Section \ref{sec:classifier}, we first use the tokenizer of BERT to tokenize the input and apply stopword lists to clean the text, thereby enhancing the model's representational capacity \cite{devlin2019bert}. Then, we construct a lightweight neural network consisting of an embedding layer, a Mamba sequence modeling layer, an adaptive pooling layer, and a fully connected layer to extract features from the processed text and perform classification. During training, we adopt the cross-entropy loss function and used AdamW optimizer to update the model parameters. We use 80\% of the data for training and the remaining 20\% for validation.
	
	During the testing phase, we apply the same preprocessing steps to the test data and evaluate it using the trained model. The corresponding experimental results are presented below.
	
	\paragraph{Experimental Validation of Prompt Classifier Effectiveness.}
	
	We compare the average accepted length across different tasks using two strategies: the prompt classifier and random assignment, where random assignment refers to randomly selecting a draft model for decoding for each input. We select 1,000 samples from each task for experimentation and experimental results demonstrate that the prompt classifier-based strategy achieves a 23\% enhancement in average accepted length when $\gamma$=10. This result reveals the optimization potential of classifier-based approaches.
	
	\paragraph{Model Architecture Comparison Experiment}
	
	We compare the model architecture proposed in this paper (Mamba-based) with the following architectures:
	\begin{itemize}
		\item MLP-based Architecture: This network architecture primarily consists of an embedding layer, two fully connected layers, and a softmax activation.
		\item CNN-based Architecture: This network architecture primarily consists of an embedding layer, convolutional layers, ReLU activation, pooling layers, and fully connected layers.
		\item BERT-based Architecture: This architecture fine-tunes the pre-trained bert-base-chinese model and adds a fully connected layer. \cite{devlin2019bert}
	\end{itemize}
	\vspace{-1mm} 
	
	We conducted a comprehensive comparison of the model architectures in terms of prediction accuracy, inference time, and model size. The experimental results are presented in Table~\ref{classifer}.
	
	Experimental results demonstrate that the prompt classifier architecture proposed in this work achieves exceptionally high prediction accuracy while exhibiting notable inference efficiency and compact model size. Specifically, the architecture attains a prediction accuracy of 99.50\% with an inference time of only approximately 0.5 seconds, averaging 0.001 seconds per item, and a model size of merely 13.49MB, making it highly suitable for deployment in resource-constrained environments. In contrast, although "BERT-based" achieves slightly higher prediction accuracy, the improvement is marginal. Moreover, its inference time reaches 354.42 seconds—several hundred times longer than that of the Mamba-based model—and its size approaches 400MB, significantly exceeding that of most baseline models used in our experiments (e.g., a draft model in this study is only 68MB). These factors impose a substantial burden for practical deployment. In summary, the proposed architecture effectively balances high predictive performance with efficiency and lightweight design, demonstrating strong potential for real-world applications.
	
	\begin{table}[ht]
		\centering
		\caption{Ablation study results for different model architectures.}
		\label{classifer}
		\begin{tabularx}{\columnwidth}{l p{1cm} p{1cm} p{1cm} p{1cm}}
			\toprule
			\textbf{Metric} & \textbf{Mamba} & \textbf{MLP} & \textbf{CNN} & \textbf{BERT} \\
			\midrule
			Prediction accuracy (\%) & 99.50  & 99.25 & 99.50  & \underline{99.99} \\
			Inference time (s) & \underline{0.54}  & 0.84  & 0.81  & 354.42  \\
			Model size (MB) & 13.49  & 17.58  & \underline{9.74}  & 390.21  \\
			\bottomrule
		\end{tabularx}
	\end{table}
	
	\section{Related Work}
	
	Since the acceptance rate and the inference efficiency of draft models are two critical factors influencing the performance of speculative decoding, we introduce relevant work on LLM speculative decoding optimization from these two perspectives.
	
	\subsection{Acceptance rate Improvement of Draft Models}
	
	Some works enhance the acceptance rate of draft tokens by increasing the number of candidate draft tokens. GSD improves the probability of draft token acceptance by generating multiple draft token branches \cite{gong-etal-2024-graph}. Specinfer treats the target model as a token tree verifier rather than an incremental decoder, improving accuracy through collective fine-tuning \cite{Miao_2024}. Similarly, Sequoia employs tree structure, introducing a dynamic programming algorithm to find the optimal tree structure for speculated tokens \cite{DBLP:journals/corr/abs-2402-12374}.
	
	In addition, some works improve the acceptance rate of draft models by dynamically adjusting $\gamma$. Work \cite{DBLP:conf/nips/KimMMMMGK23} uses fallback and rollback policies to collaboratively employ draft and target models. SpecDec++ formulates the best $\gamma$ as a Markov Decision Process (MDP) and dynamically determines $\gamma$ using a threshold policy, enhancing the inference efficiency of LLMs \cite{DBLP:journals/corr/abs-2405-19715}. Work \cite{10.5555/3692070.3692535} leverages the KV cache of the target model and the confidence scores of the draft model to enhance speculative decoding performance. DISCO dynamically modifies the number of tokens generated by the draft model during each iteration \cite{mamou2024dynamicspeculationlookaheadaccelerates}.
	
	Other works focus on optimizing the draft model to improve acceptance rate. Yang et al. propose LLMA, which utilizes the overlap between the reference and the text generation \cite{yang2023inferencereferencelosslessacceleration}. A text span is selected from the reference, and its tokens are copied into the decoder, followed by parallel verification during decoding. Online speculative decoding continuously updates the draft model with user query data observed from LLM online services \cite{10.5555/3692070.3693328}. Minions uses multiple small speculative models (SSMs) to jointly speculate the output of LLMs, enhancing acceptance rate and thus improving inference performance \cite{DBLP:journals/corr/abs-2402-15678}. The work in suggests replacing draft tokens with \(n\)-gram generated by the target model \cite{10.5555/3692070.3692631}. Observing that in speculative decoding, the acceptance probability of different tokens decreases exponentially as the distance from the first token increases, work suggests using draft models with different parameter scales at different positions in the generated text \cite{chen2024cascadespeculativedraftingfaster}. DistillSpec uses knowledge distillation to better align the draft model with the target model, and SpecDiff uses discrete diffusion models as draft models to generate draft tokens \cite{zhou2024distillspecimprovingspeculativedecoding, DBLP:journals/corr/abs-2408-05636}. Compared to DistillSpec and SpecDiff, the method proposed in this paper demonstrates stronger transferability, allowing for flexible adaptation across different domains and tasks, showcasing excellent generalizability and practicality. These approaches highlight the growing trend of utilizing tree-like structures and advanced algorithms to refine token generation and enhance decoding accuracy, a direction that our work further explores and optimizes with a focus on task-specific adaptation.
	
	\subsection{Inference Efficiency Improvement of Draft Models}
	To improve the inference efficiency of draft models, He et al. propose generating draft tokens through retrieval \cite{he-etal-2024-rest}. Staged Speculative Decoding introduces a “nested” mechanism for speculative decoding within the draft model's decoding \cite{spector2023acceleratingllminferencestaged}. At the same time, considering that loading multiple models on a single system may lead to significant computational overhead, many studies generate draft tokens directly by using the target model itself to avoid the additional model overhead \cite{DBLP:conf/icml/CaiLGPLCD24}. Specifically, SPEED improves the efficiency of decoding by predicting draft tokens using the hidden states of previous layers combined with the current token information \cite{hooper2024speedspeculativepipelinedexecution}. Additionally, another work proposes a method that selectively skips certain computational layers of the target model to accelerate the draft token generation process \cite{zhang-etal-2024-draft}. Kangaroo uses a fixed shallow sub-network as a self-draft model and introduces double early exiting mechanisms, significantly improving the inference speed and parameter utilization efficiency of large models \cite{DBLP:journals/corr/abs-2404-18911}. LayerSkip applies layer dropout and early exit loss during training and uses a self-speculative decoding scheme during inference, significantly improving inference performance while reducing memory usage \cite{DBLP:conf/acl/ElhoushiSLHWL0A24}. ReDrafter uses a recurrent neural network (RNN) as the draft model and employs a dynamic tree attention algorithm while training with knowledge distillation \cite{DBLP:journals/corr/abs-2403-09919}.
	
	In addition, some works optimize speculative decoding from other perspectives. Spectr proposes a valid draft selection algorithm whose acceptance probability is $(1 - \frac{1}{e})$-optimal multiplicatively where \(e\) is the mathematical constant \cite{10.5555/3666122.3667436}. Clover transmits the sequential knowledge of pre-speculated tokens through a regressive connection and then integrates these speculated tokens using an attention decoder, introducing an augmenting block \cite{DBLP:journals/corr/abs-2405-00263}. SLiM introduces hypothesis reduction based on fast posterior estimation, achieving significant computational savings during the verification stage \cite{DBLP:conf/naacl/LinTSHSJ24}. MagicDec uses a draft model with sparse KV cache to solve the KV bottleneck problem that increases with sequence length and batch size \cite{DBLP:journals/corr/abs-2408-11049}.
	
	Different from these existing studies, the optimization method proposes in this paper is tailored for downstream tasks. Given that draft models with limited parameters face challenges in balancing acceptance rate and inference efficiency among different tasks, fine-tuning draft models for downstream tasks can substantially enhance acceptance rate without increasing the inference overhead.
	
	\section{Conclusion}
	
	In this work, we introduce TaskSpec, a LLM speculative decoding optimization method tailored for downstream tasks. The proposed method, TaskSpec, achieves full automation in the construction of heterogeneous draft models and constructs datasets using the results aligned by the target model and trains heterogeneous draft models for different downstream tasks. Additionally, a light weight prompt classifier is designed to select the most suitable draft model for each user input. Experimental results show that TaskSpec significantly improves draft models' acceptance rate for downstream tasks, achieving a speedup up to 2.64$\times$ over the original speculative decoding. Our solution improves the acceptance rate of draft models across various downstream tasks, thereby reducing the total inference overhead of speculative decoding.
	
	\begin{acks}
		To Robert, for the bagels and explaining CMYK and color spaces.
	\end{acks}
	
	\bibliographystyle{ACM-Reference-Format}
	\bibliography{arxiv}

\end{document}